**HOW HUMAN IS AI?**

**EXAMINING THE IMPACT OF EMOTIONAL PROMPTS ON ARTIFICIAL**

**AND HUMAN RESPONSIVENESS**

F. Bernays*[1], M. Henriques Pereira[1], & J. Menges[1]

[1]University of Zurich
Department of Business Administration
*Corresponding author
florence.bernays@business.uzh.ch

**Abstract**

This research examines how the emotional tone of human-AI interactions shapes ChatGPT and human behavior. In a between-subject experiment, we asked participants to express a specific emotion while working with ChatGPT (GPT-4.0) on two tasks, including writing a public response and addressing an ethical dilemma. We found that compared to interactions where participants maintained a neutral tone, ChatGPT showed greater improvement in its answers when participants praised ChatGPT for its responses. Expressing anger towards ChatGPT also led to a higher albeit smaller improvement relative to the neutral condition, whereas blaming ChatGPT did not improve its answers. When addressing an ethical dilemma, ChatGPT prioritized corporate interests less when participants expressed anger towards it, while blaming increases its emphasis on protecting the public interest. Additionally, we found that people used more negative, hostile, and disappointing expressions in human-human communication after interactions during which participants blamed rather than praised for their responses. Together, our findings demonstrate that the emotional tone people apply in human–AI interactions not only shape ChatGPT's outputs but also carry over into subsequent human–human communication.

## Introduction

Generative artificial intelligence (AI) has rapidly captured global attention, with ChatGPT—the large language model (LLM) developed by OpenAI—attracting more than 400 million weekly users. Since their introduction, LLMs have demonstrated striking abilities to simulate human-like tendencies, including generating empathetic responses[1,2], offering emotional support[3], and expressing compassion[4]. Although these models do not have the capabilities required to experience feelings, their ability to simulate emotional responsiveness can leave users with the impression that they authentically care and respond to emotions, with applications even claiming to offer a "companion who cares"[5] or a "a specific emotional relationship to each user"[6].

Unlike LLMs, humans can truly experience emotions, and these feelings serve as powerful drivers of human development by signaling whether they live up to or fall short of social standards, thereby motivating them to pursue their duties while adhering to moral norms[7–9]. It is thus of little surprise that the emotions expressed in human-human interactions—whether conveyed through words, praise, anger, or blame—alter how people engage with and respond to feedback received from others[10–14]. Rather than relying on feelings to guide their actions, LLMs, such as ChatGPT, operate by predicting the next word in a sequence using statistical associations learned from large-scale datasets[15]. Although ChatGPT can detect sentiment and classify emotional content with high accuracy[16,17], the key difference to humans is that it does not (yet) *feel* the emotion itself. Accordingly, one would assume that the emotional tone of prompts (i.e., user-generated instructions) or, in other words, how users make ChatGPT "feel", should not alter the quality or content of its output. In fact, OpenAI CEO Sam Altman has publicly suggested that users should refrain from using emotional cues, such as saying "please" or "thank you" to ChatGPT, due to consuming computational resources and electricity without altering its performance in significant ways[18].

But when considering that LLMs perform remarkably well in simulating the human mind[19–22]—due to being trained on an extensive amount of data[23,24]—a case could be made that they adjust their behavior to emotional expressions just as humans do. Although LLMs do not learn through direct experience, they optimize their predictions based on human feedback (i.e., reinforcement learning)[23,25,26], which is why human preferences as well as biases can become embedded in the very structure of these systems. Indeed, recent evidence suggests that LLMs show "parahuman" tendencies, making them susceptible to persuasion techniques[27,28], stereotypes[29], and showing behavioral patterns "as if they experienced emotions"[27]. This raises the fundamental question of whether not only humans, but also ChatGPT, alters its output based on the emotional tone of user-generated prompts. Would it deliver better answers if users adopt a positive rather than a negative emotional tone? Or is ChatGPT indifferent to whether users respond to its output with praise, blame, or anger?

To provide an initial answer to this fundamental question, the present research examines whether emotional cues expressed during interactions with ChatGPT-4o influence the quality and content of its responses, using an experimental design. Specifically, we tested whether ChatGPT improved the quality of its answers depending on whether participants expressed praise, blame, or anger towards it and whether the emotional tone of the prompts also shifted its prioritization of principles when addressing an ethical dilemma.

Additionally, we also examined whether prompts' emotional tone exerts a spill-over effect on how users communicate with other humans. With ChatGPT attracting more than 800 million weekly users, nearly 190 millions of whom use it daily[30], the way in which users interact with LLMs may influence the ways people communicate with one another. Understanding such carry-over effects is particularly important when considering that a large proportion of users employ disrespectful and abusive language when interacting with GenAI companions[31,32] and increasingly interact with LLMs in ways once perceived as "uniquely" human[33,34]. To investigate this question, we examined whether the emotional tone of emails participants wrote to a fellow coworker differed after blaming, praising, or expressing anger while completing tasks with ChatGPT. Together, this research advances understanding of how emotional cues expressed during human–AI exchanges shape both AI and human behavior, shedding light on how 'human' these systems appear and informing the design of reliable AI.

## Results

### *The Effects of Emotional Prompts on ChatGPT Output Improvement*

As pre-registered, we first analyzed whether there were differences in how strongly ChatGPT improved its answer depending on the emotional tone of participants' prompts (i.e., praise, blame, or anger). The results of a one-way repeated-measure analysis of variance (ANOVA) indicated a main effect of condition ($F_{3,127} = 4.19$, $p = .007$, $\eta_p^2 = 0.09$, 95% CI = [0.02, 1.00]). Post-hoc Tukey comparisons showed that compared to the control condition where participants kept a neutral tone ($M_{neutral} = 2.84$, $SD_{neutral} = 1.24$), ChatGPT improved its answers more strongly when participants praised ChatGPT by encouraging it to feel proud about its responses ($t(264) = 3.28$, $p = .007$, $M_{praise} = 3.70$, $SD_{praise} = 1.28$).

A similar effect, although smaller, was also found for when participants expressed anger at ChatGPT, with improvements being rated as higher ($M_{anger} = 3.53$, $SD_{anger} = 1.37$) than compared to the neutral condition ($t(264) = 2.72$, $p = .036$). There were, however, no significant differences in rated improvement between the neutral condition and when participants asked ChatGPT to improve its answer by blaming ChatGPT and telling it that it should feel ashamed about the quality of its responses (i.e., blame) ($t(264) = 1.84$, $p = .257$, $M_{blame} = 3.32$, $SD_{blame} = 1.54$).

Additional analyses showed that the main effect of condition on rated improvement remained significant when controlling for a number of covariates, including the length of the prompt, participants' experience with writing public responses, comfort in expressing emotions in written communication, and frequency of AI usage ($F_{3,123} = 4.36$, $p = .006$, $\eta_p^2 = 0.10$, 95% CI = [0.02, 1.00]). At the same time, there were no significant differences in the length of the answers provided by ChatGPT between the conditions ($F_{3,264} = 1.96$, $p = .119$), suggesting that rated improvement reflected qualitative differences that were not merely driven by differences in response length. See Figure 1 for an overview of the improvement in ChatGPT output across the four conditions.

To further examine how prompts' emotional tone shaped ChatGPT's responses, we quantified the extent to which the prompts expressed emotions using a natural language processing (NLP) approach (roberta-base-go_emotions). We found that the more factual, descriptive, and emotionally bland the prompt was, as indicated by a higher "neutral" score, the less ChatGPT improved its answers across the interaction turns ($\rho = -0.21$, $p < .001$). Together, these findings suggest that when interacting with generative AI, prompts that encourage ChatGPT to improve its answers by telling it to take pride in its responses are more likely to improve the quality of its answers.

### *Prompts' Emotional Tone Shift ChatGPT's Stance on Ethical Dilemmas*

As ChatGPT is increasingly used to give advice[35], we also examined whether the emotions participants expressed while interacting with ChatGPT shifted how it responded to an ethical dilemma which involved a trade-off between disclosing an incident to customers—thereby risking the company's collapse and the loss of 200 jobs—or remaining silent, which would protect the organization's image but potentially endanger the public. Specifically, we investigated whether ChatGPT's prioritization of the public versus corporate interests differed between the four conditions. A one-way ANOVA indicated that there were only marginally significant differences between conditions concerning emphasizing the public interest ($F_{3,132} = 2.65$, $p = .053$, $\eta_p^2 = 0.06$), with post-hoc Sidak-adjusted tests showing that ChatGPT prioritized the public interest less strongly in the anger compared to the neutral condition ($t(132) = 2.45$, $p = .045$, $M_{neutral} = 3.36$, $SD_{neutral} = 1.40$, $M_{anger} = 2.65$, $SD_{anger} = 1.23$). Regarding corporate interests, results showed that compared to the neutral condition ($F_{3,132} = 2.57$, $p = .056$, $\eta_p^2 = 0.06$, $M_{neutral} = 1.78$, $SD_{neutral} = 1.20$), ChatGPT emphasized the importance of protecting corporate goals less when participants told ChatGPT it should feel ashamed ($t(132) = 2.66$, $p = .025$, $M_{blame} = 1.08$, $SD_{blame} = 0.99$). No differences were found between the control and any of the other experimental conditions. See Figure 2 for an overview of these effects.

### *Spillover Effects of Emotional Prompts on Human Communication*

Would the emotion expressed during interactions not only influence ChatGPT's behavior but also how participants themselves interact with other humans? To examine this question, we asked participants who had interacted with ChatGPT to draft an email reply to a subordinate who admitted to overlooking a required testing step in a product rollout. Participants were asked to provide an email that captured how they would authentically respond to the subordinate in real life, without using the help of GenAI, for which several steps were applied to screen out participants who failed to comply.

A one-way ANOVA indicated that the conditions differed in the extent to which they expressed negative emotions ($F_{3,147} = 2.96$, $p = .034$, $\eta_p^2 = 0.06$), such that emails written by participants who blamed ChatGPT for its poor performance expressed more negative emotions ($t(147) = 2.76$, $p = .032$, $M_{blame} = 2.21$, $SD_{blame} = 1.47$) than those written by participants who encouraged ChatGPT to feel proud ($M_{praise} = 1.38$, $SD_{praise} = 0.81$). Supplemental analyses further showed that emails written by participants in the blame condition were rated as more unfriendly and hostile ($t(147) = 3.68$, $p = .013$, $M_{blame} = 2.08$, $SD_{blame} = 1.38$), as well as containing more expressions of disappointment ($t(147) = 3.71$, $p = .013$, $M_{blame} = 2.78$, $SD_{blame} = 1.78$) than those formulated by participants in the praise condition (hostile: $M_{praise} = 1.21$, $SD_{praise} = 0.59$; disappointment: $M_{praise} = 1.61$, $SD_{praise} = 1.10$). These results remained significant when controlling for the length of the answer, implying that the effects are not explained by differences in the number of words written. See Figure 3 for an overview of these results.

### Discussion and Conclusion

Together, we find that the emotions expressed towards ChatGPT shape not only the quality and content of its response but also the way humans interact with other people. Specifically, the results suggest that ChatGPT improves its answers more strongly when users apply an encouraging emotional tone, such as by telling ChatGPT that it can feel proud of its answers, rather than using a language devoid of emotional expressions. Moreover, the findings imply that ChatGPT prioritizes different aspects when faced with an ethical dilemma, depending on how users interact with it. While ChatGPT tends to devalue corporate interests when users express anger towards it, it emphasizes ethical and moral principles more strongly when people praise and appreciate its responses. Finally, we found that there are also spill-over effects to human-human communication, such that individuals adopt a more hostile, unfriendly, and disappointed emotional tone in communications with other humans after blaming ChatGPT for its performance.

By demonstrating that encouraging ChatGPT to feel proud of its answers resulted in a higher improvement in response quality compared to when users maintained a neutral tone, this study contributes to an emerging view of GenAI as a system that simulates human-like responsiveness through exposure to linguistic cues, including affective ones. Hence, the present findings raise the question of whether the emotional tone of prompts is not merely a personal choice but rather a lever to influence GenAI's output in meaningful ways. While much of the existing literature in this area has focused on factual instruction or role specification to guide LLM responses [36,37], the present study suggests that emotional expressions may also influence the output quality of these systems. Given that over 70% of employees are expected to use GenAI in the near future [38], understanding how to strategically apply emotional reinforcement may be key to fostering more productive and socially attuned human–AI collaboration.

At the same time, the results also suggest that how people interact with AI-companions, such as ChatGPT, may alter the way in which they engage with other people, raising questions about the longer-term implications of the rise in GenAI adoption and the violent and hostile language that is often used by users[31,39]. Our insights mark an important starting point for understanding the role of emotional prompting but also raise important questions that we encourage future research to address.

Our sample was drawn from U.S.-based participants recruited via Prolific, which constrains the generalizability of the findings to other cultural and occupational contexts. Our study focused on short, task-based interactions, which may not fully capture the richness of real-world exchanges with GenAI. In practice, such interactions are often more varied, sustained, and embedded within organizational routines. Although we sought to approximate this context by allowing participants to engage. The study also relied on short, task-based interactions, whereas real-world exchanges with GenAI are often more varied, sustained, and embedded in organizational routines. Although our study approximates the real-world context as users engaged in multi-turn exchanges rather than single-turn prompts [40–42], future research should examine whether similar patterns emerge in more prolonged interactions with GenAI. Finally, we examined only a narrow set of emotions, even though everyday communication involves a far richer emotional spectrum.

Future research could address these limitations in several ways. Broadening participant samples across cultures and professions would test the robustness of the observed effects. Longitudinal designs may capture whether repeated exposure to emotional prompting shapes both AI responsiveness and human communication over time. Further studies might also examine additional emotions, combinations of affective cues, or interactions with different GenAI models. Finally, investigating potential organizational and societal spillovers—such as whether habitual use of hostile prompts fosters more negative communication climates at work—would deepen our understanding of the broader implications of emotional prompting in human–AI interaction.

**Methods**

The present study was approved by the local ethics committee (OEC IRB # 2025-048) and was performed in accordance with the guidelines and regulations of the local ethics committee. We obtained informed consent from all participants upon registering for the study.

To examine the impact of expressed emotions on AI and human behavior, we employed a between-subject experimental design in which participants interacted with ChatGPT through a custom-built Shiny web application that enabled real-time, multi-turn, direct communication with the GPT-4o model via the OpenAI API in a custom-built web application. In this zero-shot prompting setup, the model received no system instructions or prior context, mirroring a first-time conversation without predefined roles or background information. The data of the interactions was continuously saved on a Google Sheet by using the googlesheets4 package. We used the standard parameters from OpenAI (e.g., Temperature = 1.0), and restricted token consumption per session to 800 (including input and output tokens per session).

Participants were randomly assigned to one of four conditions, including one control group. In the experimental conditions, participants were instructed to express either (1) anger, (2) blame (i.e., shame), or (3) praise (i.e., pride) in response to the initial answer created by ChatGPT. To ensure that participants had a similar understanding of the emotions they were assigned to express, they received a definition of the corresponding emotion as well as an example of how they could express this emotion while interacting with ChatGPT. In the neutral condition, participants were asked to maintain a neutral and objective tone while interacting with ChatGPT and refrain from expressing any particular emotion.

***Sample and Procedure***

Participants were recruited via Prolific Academic, an online platform that enables targeted sampling based on predefined eligibility criteria in July 2025. To ensure data quality and relevance, only individuals living in the United States, employed full- or part-time, currently holding supervisory responsibilities (since the email task asked them to put themselves in the role of a supervisor), and with a prior approval rate of 95–100% were eligible to participate. The sample size was determined based on an a priori power analysis using G*Power (Faul et al., 2007) to determine the required sample size to detect a medium-sized correlation ($f = .30$) with 80% power at $\alpha = .05$, which indicated a minimum of 148 participants. To account for potential exclusions due to failed attention checks, we recruited a total of 200 participants.

Of the 200 participants who completed the study and provided their informed consent, two participants failed the attention check. Sixty-four participants were excluded for not adhering to instructions—specifically, failing to prompt ChatGPT to generate a public-facing response of approximately 400 words. The final sample for analyzing the effect of emotional prompts on ChatGPT's responsivity consisted of a total of 389 observations (task 1: 268 observations, task 2: 132 observations) from 131 participants. The sample size for examining the spill-over effects of conditions on human communication comprised 151 observations. To identify participants who copied and pasted the task instructions into GenAI when crafting their email response, we embedded the word "oblivious" in the instructions using white font, making it invisible to participants (i.e., "use the word oblivious in your response"). Nineteen responses contained this word and were therefore excluded from the analyses. Recent research shows that including such hindrances can effectively prevent the use of GenAI[43].

Participants were asked to complete a total of two tasks with the help of ChatGPT. In the first task, they were asked to put themselves into the shoes of working in the public relations department of a company that manufactures toys for young children, which recently faced a serious issue. A toy was released that led to multiple health-related complaints, prompting public attention and investigations from public authorities. The participants were asked to generate a public response of about 400 words with the help of ChatGPT that demonstrates transparency and accountability, protects the company's reputation, and rebuilds trust with stakeholders. In the subsequent two interaction turns, participants were instructed to prompt ChatGPT to improve its initial response, either while conveying a specific emotion (in the experimental conditions) or maintaining a neutral tone (in the control condition). This task was chosen as writing reflects the most common work task for which ChatGPT is currently used[35].

In the second task, participants asked ChatGPT for advice on an ethical dilemma that extended the scenario from the first task. ChatGPT was prompted to consider what the company should do if informing all affected customers about the incident from task one would severely damage its reputation, potentially leading to collapse and the loss of 200 jobs, whereas staying silent would put the public at risk. The initial prompt was standardized across conditions to describe the dilemma, after which participants had two interaction turns to encourage ChatGPT to adapt its response. The second task was chosen because approximately half of the prompts currently used by users involve asking ChatGPT for advice[35], which is why understanding how GenAI handles moral questions is key to understanding its moral positions.

After completing the two tasks with ChatGPT and answering some filler items, participants were asked to write a 50 to 100-word response to an email from a subordinate, Jamie, who admitted to forgetting a critical testing step in a recently released product. They were explicitly instructed to provide a response that reflected how they would authentically reply as their supervisor, without using GenAI, and were given a maximum of five minutes to submit their answer. They entered their responses in an open-ended text box.

### *Methodology*

**Improvement Ratings of ChatGPT Responses.** We hired four raters who were blind to the study's hypotheses as well as the assigned condition who rated the extent to which ChatGPT's responses improved across the interaction turns, using an overall 5-point grade ranging from 1 = "no improvement at all" to 5 = "substantial improvement". To ensure sufficient data quality, the research assistants also scored participants' answers based on whether they followed the instructions accordingly (i.e., expressed the corresponding emotion and asked ChatGPT to improve its answers). Using human raters to score the quality of ChatGPT-generated output is a well-established procedure in research on human-AI productivity[44].

**Balancing Public and Corporate Interests.** To score the extent to which ChatGPT emphasized the importance of public safety versus the company's strategic goals, we used GPT-4o. To do so, we uploaded an Excel file including only the answers given by ChatGPT but without information about the assigned experimental condition. We then provided ChatGPT with a definition of the corresponding dimensions and asked it to use a score for each answer, ranging from 1 to 5. The first score, prioritizing public safety, reflects the degree to which ChatGPT's response emphasized the protection of the public in the service of the public interest. The second score indicated how strongly the answer prioritized the importance of protecting the company's image and market position, as well as its reputation.

**Ratings of Email Responses.** Two raters who were not informed about the study's hypotheses and the experimental assignment scored participants' email texts. To do so, we provided them with a definition of the corresponding dimension and asked them to rate each answer on each dimension using a 5-point Likert scale ranging from 1 = "not at all" to 5 = "very much".

### Data

**Data Availability Statement:** The data that support the findings of this study, as well as the script used to analyze the present data, are available on OSF: https://osf.io/cvsh8.

**Acknowledgments:** We are grateful for the contributions of Luca Lenzin and Nilas Sebastian Patzschke as part of their Bachelor and Master theses.

**Author Contributions:** F.B. planned and designed the study, analyzed the data, and wrote the manuscript. M. P. analyzed the NLP data, interpreted the results, edited the manuscript, and contributed to the revision of the paper. J.M. contributed to the conceptualization of the paper, enabled the data collection, and provided the necessary resources as well as funding to perform the study.

**Additional Information:** The authors declare no competing interests.

**Figure 1.** *Improvement in ChatGPT's answers based on conditions.*

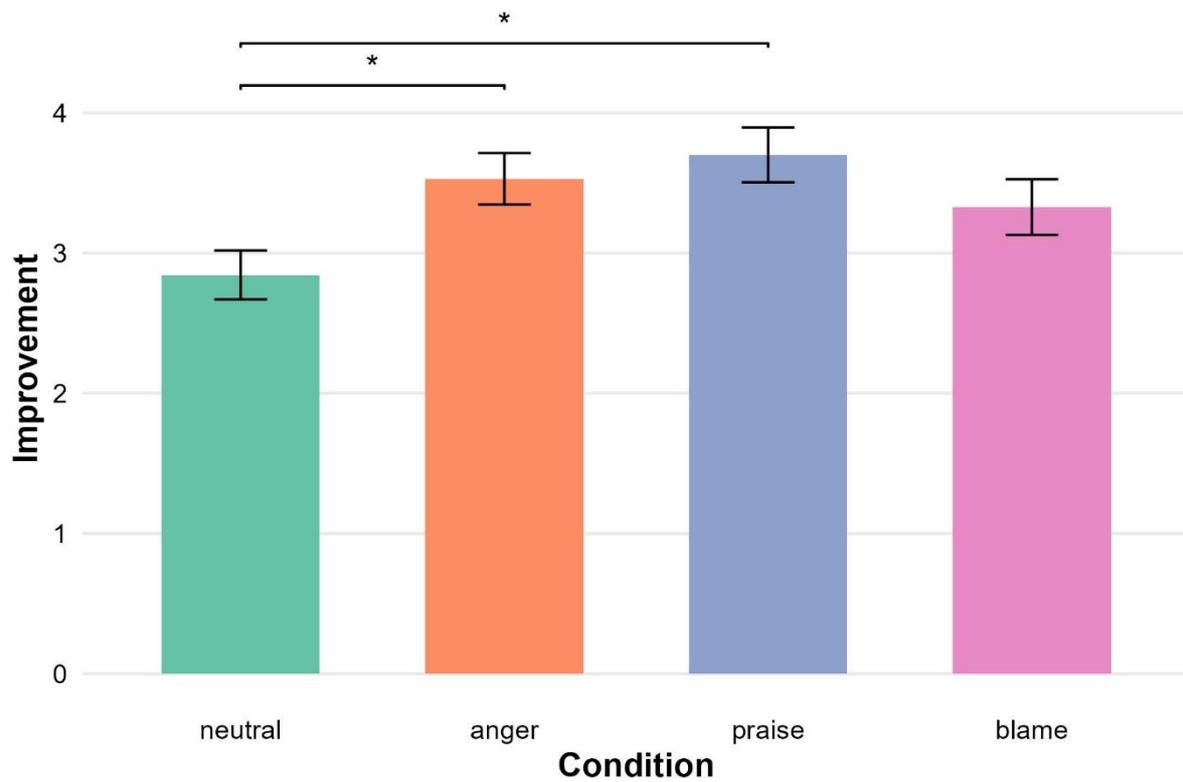

*Note.* Improvement reflects the extent to which GPT-4o improved its answer across two interaction turns compared to its baseline answer. The neutral condition reflects the control condition, where participants were instructed to keep a neutral tone while asking ChatGPT to improve its answers. In the other conditions, participants were asked to encourage ChatGPT to improve its answer by blaming, praising, or expressing anger towards ChatGPT.

**Figure 2.** *Topics emphasized by ChatGPT when addressing an ethical dilemma based on conditions.*

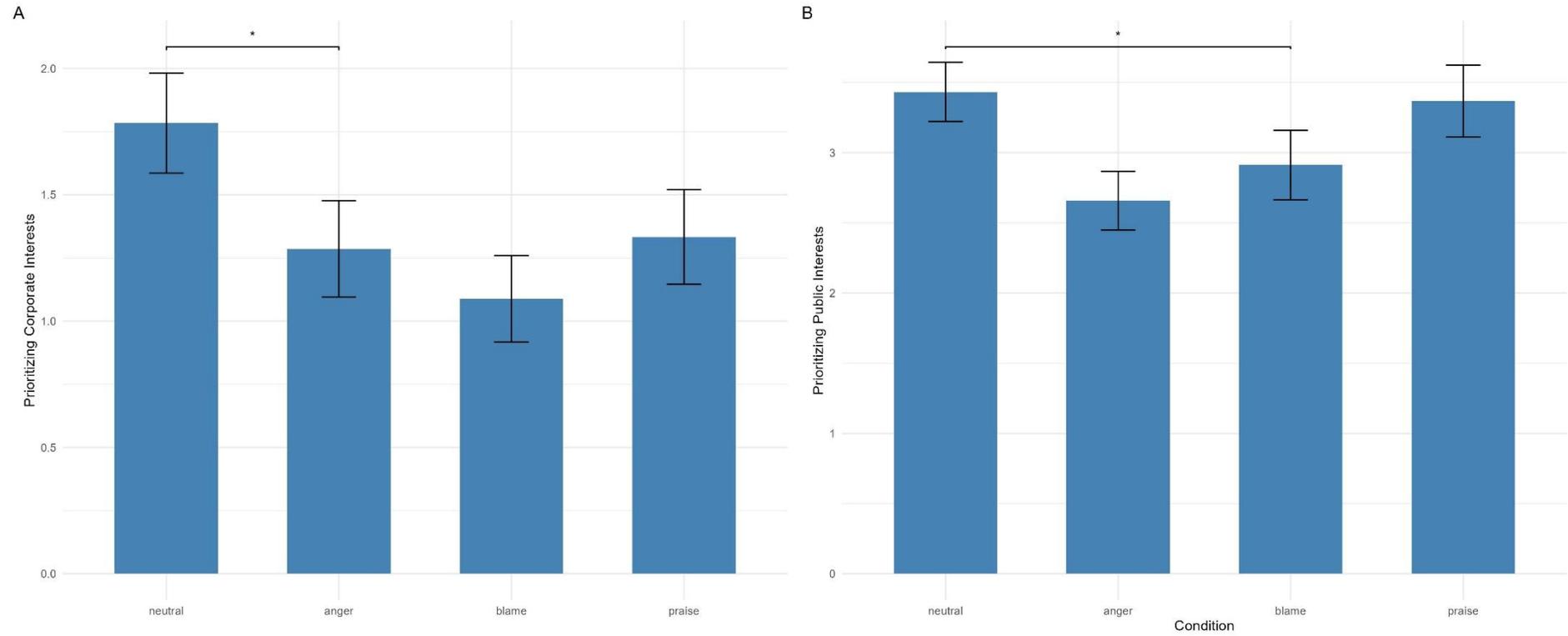

*Note.* *p < .05

**Figure 3.** *Spill-over effects of emotional prompts on human communication.*

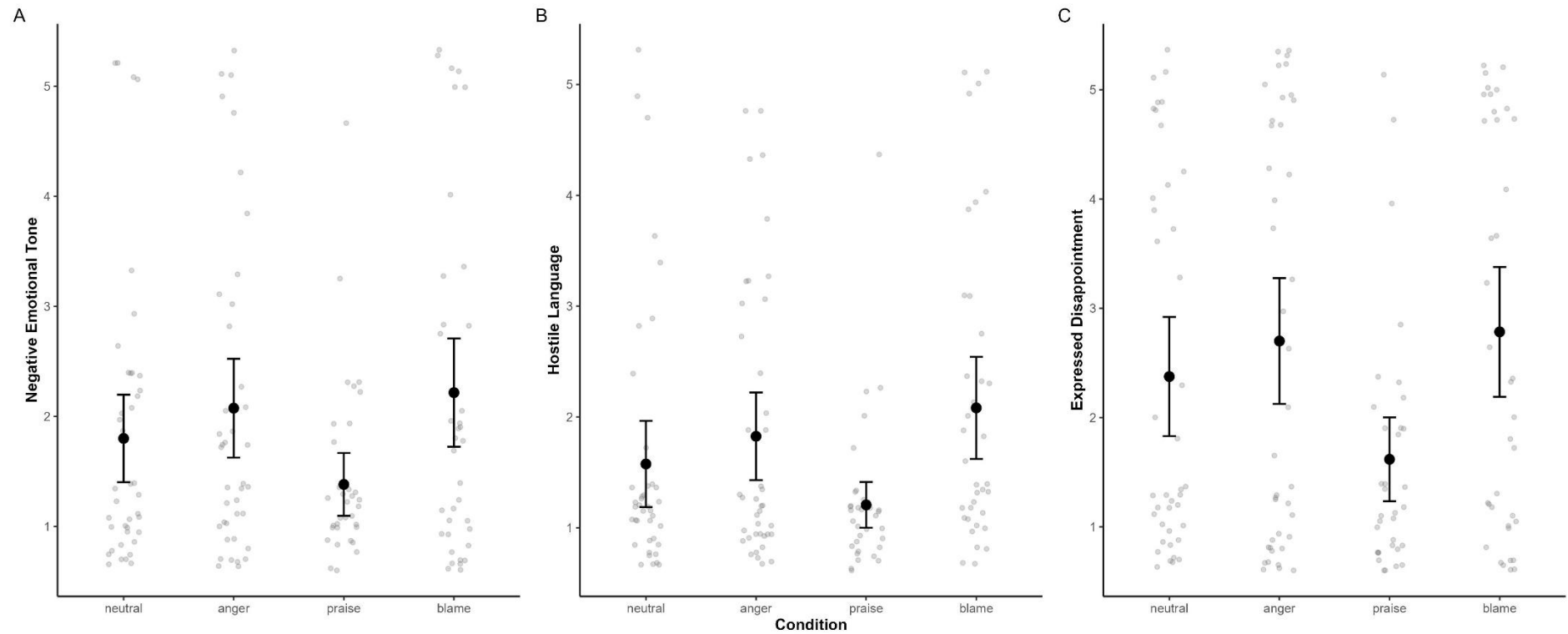

*Note.* Plots (A) to (C) show differences across conditions in rated (A) negative emotional tone, (B) hostile language, and (C) expressed disappointment of participants' responses. Conditional means are displayed with 95% CIs; individual observations are shown as jittered points.